# Self-supervised monocular depth estimation from oblique UAV videos

Logambal Madhuanand, Francesco Nex, Michael Ying Yang

Faculty of Geo-Information Science and Earth Observation (ITC), University of Twente,
Enschede, The Netherlands
logambal.ceg@gmail.com, f.nex@utwente.nl, michael.yang@utwente.nl

**ABSTRACT:**

UAVs have become an essential photogrammetric measurement as they are affordable, easily accessible and versatile. Aerial images captured from UAVs have applications in small and large scale texture mapping, 3D modelling, object detection tasks, DTM and DSM generation etc. Photogrammetric techniques are routinely used for 3D reconstruction from UAV images where multiple images of the same scene are acquired. Developments in computer vision and deep learning techniques have made Single Image Depth Estimation (SIDE) a field of intense research. Using SIDE techniques on UAV images can overcome the need for multiple images for 3D reconstruction. This paper aims to estimate depth from a single UAV aerial image using deep learning. We follow a self-supervised learning approach, Self-Supervised Monocular Depth Estimation (SMDE), which does not need ground truth depth or any extra information other than images for learning to estimate depth. Monocular video frames are used for training the deep learning model which learns depth and pose information jointly through two different networks, one each for depth and pose. The predicted depth and pose are used to reconstruct one image from the viewpoint of another image utilising the temporal information from videos. We propose a novel architecture with two 2D CNN encoders and a 3D CNN decoder for extracting information from consecutive temporal frames. A contrastive loss term is introduced for improving the quality of image generation. Our experiments are carried out on the public UAVid video dataset. The experimental results demonstrate that our model outperforms the state-of-the-art methods in estimating the depths.

## 1. INTRODUCTION

Unmanned Aerial Vehicles (UAVs) are commonly used photogrammetric measurement platforms that gained popularity due to its accessibility, affordability and its versatility in capturing images. The images captured from UAVs can be used for applications like Digital Surface Model (DSM) generation, Digital Terrain Model (DTM) extraction, mapping, textured 3D models, etc. (Nex and Remondino, 2014). The acquisition of images by UAVs can either be through nadir view or oblique view based on its application. The images acquired from nadir view have a constant scale and good coverage whereas oblique view images have wider coverage, provides higher redundancy, better 3D model formation and can be also useful for an improved orthophoto generation (Nex et al., 2015). Yet, oblique images have a varying scale, occlusions and illumination problems which requires detailed processing (Aicardi et al., 2016). A set of oblique view images with sufficient overlap, acquired from UAVs can be used for reconstructing a dense 3D model through photogrammetric image techniques like Structure-from-Motion (SfM). The general steps involve feature detection and matching by finding correspondences between images, sparse reconstruction, bundle adjustment followed by 3D dense point clouds generation (Vallet et al., 2012; Voumard et al., 2018; Nex et al., 2015). However, these imaging techniques suffer from issues like requirement of multiple images of the same scene, time consumption, low-quality generation in regions where the camera has a narrow view and for images captured along camera optical axis. This led to an increased interest towards image-based 3D reconstruction techniques that can overcome these problems.

Recent advancements in computer vision and deep learning techniques have increased the focus towards depth estimation by using single images, doing away with the need for multiple images (Eigen et al., 2014; Liu et al., 2015; Garg et al., 2016; Godard et al., 2017). Single Image Depth Estimation (SIDE) models use depth cues like contextual information, texture variations, gradients, shading, defocus etc., for obtaining depth images from a single (monocular) image (Saxena et al., 2005). A depth image represents the distance from the viewpoint to the scene object at a particular orientation in each pixel. These depth images have applications in 3D reconstruction, semantic segmentation, object identification and tracking, monitoring topographic changes, etc. (Chen et al., 2018). Most of the studies that deal with single image depth, use Convolutional Neural Networks (CNN) as CNNs have proven capable of learning fine details by combining low level and high level features (Bhandare et al., 2016) that are beneficial for this extraction task. Nearly all previous works dealing with SIDE models have been focused on images taken at ground level or indoor images while monocular depth estimation from aerial images has been attempted very rarely. Compared to terrestrial images, aerial images may lack information like shading, texture etc., making it more difficult for the learning process and also its varied perspectives along with its increased distance from the camera point makes it a challenging task to be addressed. Single image depth from UAV aerial images can have many applications, such as, for preparing low cost elevation models, for 3D reconstruction with a minimal number of images captured beforehand for reconstruction, damage evaluation in places where regular photogrammetric block acquisition is not possible, on-board UAVs for augmented Simultaneous Localization And Mapping (SLAM) to estimate the position of vehicles etc. Representative monocular depth estimated from single UAV images in UAVid dataset (Lyu et al., 2020) is shown in Figure 1.

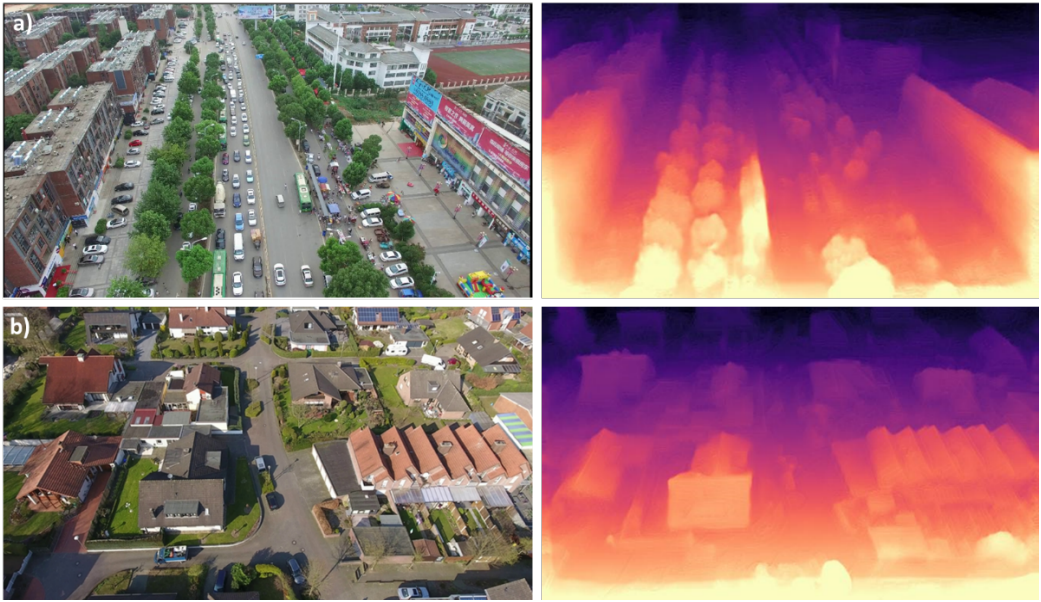

Figure 1. Disparity from two representative monocular images in UAVid dataset (Lyu et al., 2020) a)Wuhan, China, B) Gronau, Germany

Monocular depth using deep learning models have been commonly approached by training the CNN network with corresponding ground truth depth maps (Eigen et al., 2014; Liu et al., 2015; Laina et al., 2016; Li et al., 2017 ; Mou and Zhu, 2018; Koch et al., 2019; Amirkolaee and Arefi, (2019)). Even though the predicted depth maps from supervised SIDE models look very similar to the ground truth depth, collecting large quantities of pixel-wise ground truth depth for training is a tiresome process. This led towards the development of Self-Supervised Monocular Depth Estimation (SMDE) techniques. SMDE techniques learn depth from complementary information with depth results comparable in accuracy to that of supervised depth learning model. Available SMDE techniques learn depth through either stereo pairs or monocular videos. When using stereopairs, the SMDE model takes one image from the pair as input to the deep learning model to predict disparity. The predicted disparity is warped with the other image from the pair to generate the original input image. The difference between the generated and original image is backpropagated for improving the model performance. The produced disparities can be converted to depth using the baseline and focal length which is usually done as an offline process. However, preparation of stereopairs dataset from UAV aerial images is a challenging task and the model trained with stereopairs might have issues due to occlusions (Godard et al., 2017). While using monocular videos for training, the SMDE model learns both depth and pose information jointly through two different networks, one each for depth and pose. The predicted depth and pose are used to reconstruct one image from the viewpoint of another image utilising the temporal information from videos. Acquiring videos are becoming increasingly popular as they offer flexibility during data collection. With the developments in instruments, it has become possible to acquire videos at resolutions that are comparable with high quality still images. As most of the UAV platforms captures videos, depth estimation from monocular videos through SMDE is preferred.

In this paper, we propose an architectural improvement and an additional loss term for enhancing the qualitative and quantitative performance of monocular depth model estimation from oblique UAV videos. The contributions of this work are the following. (1) To the best of our knowledge, we present the first work of self-supervised monocular depth estimation model for oblique UAV videos. (2) Technically,  we use two consecutive temporal frames in two 2D CNN encoders to form feature maps that are given as an input to 3D CNN depth decoders for generating inverse depths. (3) Furthermore, we propose to add a contrastive loss term in the training phase to make the model produce images closer to the original video images.

This paper is organised as follows. Section 2 gives an outline about the related work similar to our work. Section 3 presents the structure of the proposed model architecture. The experimental implementation and the results are explained in Section 4, which is followed by conclusions.

## 2. RELATED WORK

Depth estimation from images using computer vision techniques is very popular due to its successful performance with terrestrial images. It includes the use of multiple image views of the same scene (Szeliski & Zabih, 2000; Remondino et al., 2013; Furukawa & Hernández, 2015), stereopairs (Alagoz, 2016),  illumination or texture cues (Zhang et al., 1999) etc., to reconstruct the 3D model. Extraction of 3D information from single images relied on techniques like shape from shading (Van Den Heuvel, 1998), shape from texture (Kanatani and Chou, 1989), followed by the use of stereo or temporal sequences from 2D images. Based on the additional constraints used for extracting depth from single images, they are clustered into different categories. Li et al., (2020) reviewed various SIDE models and datasets along with the applications benefitted from these models. The SIDE models explained below are some of the important studies which contributed to the development of our model and are mostly based on SIDE models that are applied to aerial/UAV images.

## 2.1 Depth from stereo images

Binocular stereo or multi-baseline stereo (Kang et al., 1999) techniques are used to extract 3D information from two or multiple images covering the same scene. This involves matching pixels across multiple rectified images and using them to orient the images for the point cloud generation. The distance between these corresponding points in the left and right image defines the disparity map of the images. This disparity map can be used for 3D reconstruction. The disparity and depth information are related inversely as $Disparity = X_l - X_r = \frac{Bf}{d}$, where $X_l$ and $X_r$ denote the corresponding image points, B represents the baseline distance between cameras, f is the camera constant and d is the depth or object distance from the viewpoint. The obtained disparity map can be used to calculate the depth information from the images through the baseline and camera constant. This process of finding the difference between corresponding points from two or more viewpoints forms the base of stereo matching algorithms. Stereo matching algorithms include estimating the matching cost, cost aggregation, disparity calculation and optimization using low level image features to measure dissimilarity (Zhou et al., 2020). One among them is Semi-Global Matching (SGM) proposed by Hirschmüller, (2007) that has wider adoption in many recent computer vision tasks due to the quality results and its faster performance. SGM is a dense image matching technique, which matches pixel-wise mutual information by matching cost. Instead of using intensity difference alone for matching, SGM uses disparity information to find the corresponding pixels in other images. Deep learning has been successfully used to overcome many of the difficulties in traditional stereo algorithms. Žbontar and Le Cun, (2015) used CNN to compute matching cost and depth from the rectified image pair. Mayer et al., (2016) used FlowNet to estimate both disparity and scene flow with large scale datasets. Kendall et al., (2017) employed 3D convolutions using GC-NET for regularizing cost volume and regressing disparity from stereo images. Liang et al., (2018) proposed a network architecture to include all steps of stereo matching algorithms by using a different disparity refinement network for joint learning. There are also other deep learning approaches which produce highly accurate depths from stereo image pairs. However, preparing large training datasets with corresponding ground truths is difficult. Many of these methods acted as a base for single image depth estimation techniques where the limitation of acquiring multiple images of the same scene can be overcome.

## 2.2 Single image Depth Estimation with Supervised Learning

Depth from single images has been in existence for a long time and has been implemented by imposing specific conditions or adding complementary information. Saxena et al., (2005) used Markov Random Fields (MRF) to combine local and global features to produce depth using the contextual relationship between features while avoiding occlusions. With recent developments in computer vision and deep learning techniques, there is an increasing possibility of overcoming the limitations of analytical methods. This is mainly due to the success of Convolutional Neural Networks (CNN) in learning depth from colour intensity images. Several studies have been published on depth estimation from a single image using ground truth depths for training deep learning models. Most of the studies have been on indoor images or outdoor images taken at ground level. Only a few studies have analysed the techniques for depth estimation from aerial images. Julian et al., (2017) compared different style transfer methods like pix2pix, cycle GAN and multi-scale deep network for aerial images captured from UAVs. They trained the model using the UAV images along with depth image pairs and refined the feature-based transfer algorithm for this single image depth estimation purpose. Mou & Zhu, (2018) used a fully residual convolutional-deconvolutional network for extracting depth from monocular imagery. They used aerial images along with the corresponding DSM generated through semi-global matching for training the network. The two parts of the networks acts as a feature extractor and height generator. Amirkolaee & Arefi, (2019) proposed a deep CNN architecture with an encoder-decoder setup for estimating height from aerial images by training them with the corresponding DSM. They extracted the full satellite image into local patches and trained the model with the corresponding depth and finally stitched the depths together. They faced issues for small objects with fewer depth variations like small vegetation, ground surfaces within the scene etc. Although all these methods proved to be successful, they all require huge amounts of ground truth depth images while training the model. Preparing large amounts of UAV images along with their corresponding DSM is complicated, making supervised approach less preferable compared to other approaches even though it produces slightly better accuracies for single image depth estimation.

## 2.3 Single Image depth estimation with Self-Supervised Learning using stereopairs

The limitation in acquiring large ground truths for training led towards the growth of studies towards SMDE techniques. The reduced dependencies on laborious ground truth data collection have generated a lot of interest in these approaches. Garg, Vijay Kumar, Carneiro, & Reid, (2016) circumvent the problem faced by supervised learning by utilising stereo images instead of ground truth depth maps. They used the 3D reconstruction concept to generate a disparity image from stereo images and reconstruct the original image through inverse warping. They suggested this approach can be continuously updated with data and fine-tuned for specific purposes. Although the model performed well, their process of generating images is not fully differentiable due to the non-linear loss terms. The use of Taylor approximation makes the loss terms more complex. Godard et al., (2017) overcame this by including a fully differential training loss term for left-right consistency of the generated disparity image to improve the quality of the generated depth image. The adversarial learning models mark the current state of the art in many areas where deep learning is being used. A simple GAN network consists of a generator that learns to produce realistic images and discriminator that learns to find the difference with real images. MonoGAN by Aleotti, Tosi, Poggi, & Mattoccia, (2018) used a combination of generator and discriminator network for the monocular depth estimation. Mehta, Sakurikar, & Narayanan, (2018) used this structured adversarial training to improve the task of image reconstruction for predicting depth images from the stereo images. Tosi et al., (2019) used traditional stereo knowledge along with the combination of synthesised depths for estimation of depth from monocular images. Madhuanand et al., (2020) used stereo images for depth estimation from single UAV images using two deep learning models and compared the model performances. These SMDE models use stereo pairs for replacing the ground truth information while training.

## 2.4 Single Image depth estimation with Self-Supervised Learning from videos

UAV videos are easier to acquire than producing stereo pairs though there are additional complexities in a UAV video-based SMDE model as it needs to estimate both depth and position. Zhou et al., (2017) proposed a novel framework to jointly learn depth and ego-motion with videos using adjacent temporal frames in a self-supervised manner. Similarly, Vijayanarasimhan et al., (2017) used SfM-Net for geometry aware depth estimation for certain objects in a scene with an option available for making it a supervised approach. Following this, different approaches for estimating depth from monocular videos have been proposed. Mahjourian et al., (2018) used 3D geometrical constraints for improving the monocular depth estimation performance. Other works which have modifications in depth estimation network or additional loss terms for increased performance includes Poggi et al., (2018), Wang et al., (2018), Godard et al., (2019), Dai et al., (2019), Tan et al., (2020), Spencer et al., (2020), Guizilini et al., (2020) etc. Closer to our work, Hermann et al., (2020) used monocular UAV videos for depth estimation. They have proposed a joint estimation of depth and pose network for simultaneous learning and prediction. They have suggested three losses, image reconstruction loss, smooth loss and edge-aware loss, for improving the performance. Their study used three different datasets, rural, urban and synthetic dataset, and evaluated the model performance separately for the datasets.

## 3. METHOD

This section describes the overall methodology of the self-supervised depth estimation model from UAV oblique videos. This involves the preparation of input images to the network, training the deep learning model and fine-tuning the model for hyper-parameters. The model predicts depth map using depth network and then estimates rotation and translation parameters from consecutive temporal frames using the pose network which are used to reconstruct the given input image. The difference between the reconstructed and the original input image is calculated as a loss to be back-propagated for improving the model performance. The quality of the estimated single depth images is evaluated using various statistics by comparing with ground truth or reference depth images produced using Pix4D ("Pix4D (version 4.4.12)," 2020). The architectural outline of the methodology is shown in Figure 2.

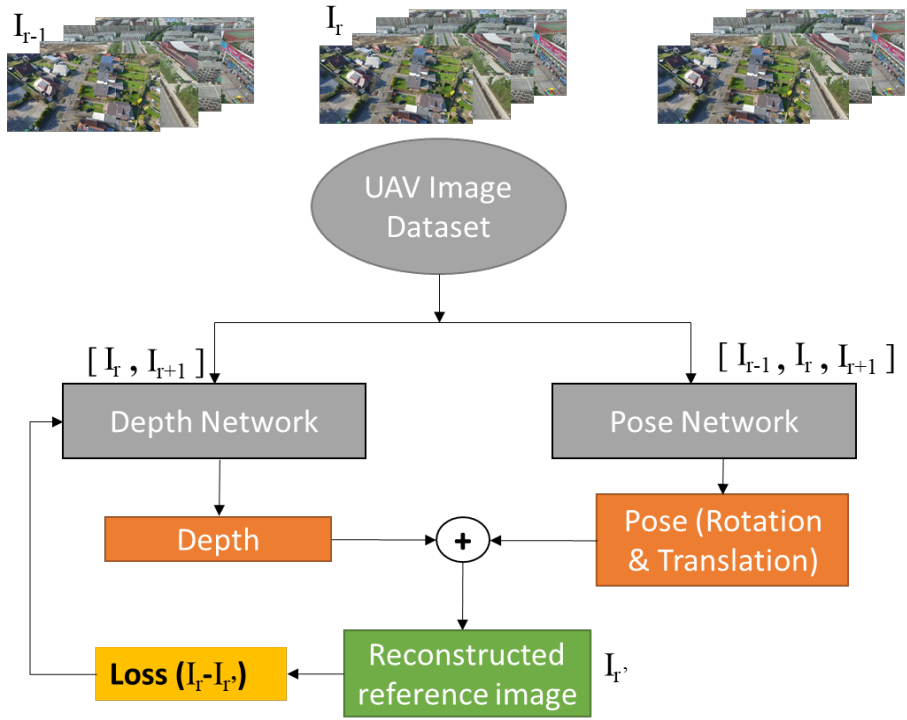

Figure 2. Network architecture with Depth network and Pose network taking input from the UAV video dataset.

The architecture includes two successive temporal frames used as an input to two 2D encoders which are extracted as depth through a 3D depth decoder. This predicted depth along with pose information using three temporal frames is used to reconstruct the input image. The difference between the reference and the original input image is used as a loss to improve the model performance.

## 3.1 Input to Network

The SMDE model requires a large amount of training images. The training images are extracted from temporal frames of UAV videos. The training involves sets of three consecutive temporal images that are provided as an input to the model. The three images are chosen such that the baseline between them is maintained without disturbing the image reconstruction process. The three temporally adjacent images are referred as $I_{r-1}$, $I_r$, $I_{r+1}$, where $I_r$ represents the reference image, $I_{r-1}$ and $I_{r+1}$ represents the source images at a previous frame and image at a subsequent frame respectively. The camera intrinsic parameters which include principal point location and focal length information are required to estimate the pose information.

## 3.2 Network Architecture

The SMDE model consists of two networks, one each for estimating depth and pose information from temporal image sequences. The depth network is used to estimate inverse disparity through the image reconstruction process. The predicted pose and depth along with the source images is used to reconstruct the reference image. This is achieved by taking Depth image (D) generated from depth network, camera transformation matrix ($P_{r \rightarrow r'}$) from pose network and a source image as input. It uses bilinear sampling by sampling pixels for projecting the source view with depth and pose information to reconstruct the reference image as shown in Equation (1).

$$I_{r' \rightarrow r} = I_{r'}[proj(D_r, P_{r \rightarrow r'}, K)] \tag{1}$$

where $I_{r' \rightarrow r}$ represent the reconstructed image by the projection of source image into reference image, $I_{r'}$ represents the source image, $D_r$ represents the predicted depth, $P_{r \rightarrow r'}$ represents the relative camera pose between the reference and source image, K represents the camera intrinsic parameters.

### 3.2.1 Depth Network

For the depth estimation task, we follow the encoder-decoder network architecture similar to the one proposed by Godard et al., (2019). In order to extract more information from the input images, we propose two encoders constructed with ResNet18 (He et al., 2016) architecture. The encoders use pre-trained weights to guide the model towards global minimum. The two encoders take as input the reference image ($I_r$) and the successive temporal source image ($I_{r+1}$). They pass through the encoder with multiple skip connection to extract both low level and high level features. Using two image frames instead of one can help in extracting more features and can also be used for building occluded objects in one frame with the other. The two images are passed through the first layer with a kernel size of 7x7, followed by 3x3 filter with an increasing number of features at each layer. The outputs from the two encoders are concatenated to form an extra dimension. The depth decoder is formed with a 3D convolutional network that uses the volumetric information from both features. This 3D CNN can help in extracting features from temporal sequences through convolutional blocks. This is followed by up-sampling layers to match the layers as that of the original image size. The features from all the layers are averaged to form the last layer which consists of sigmoid activation to predict the Depth (D) map. The network architecture is explained in detail in Figure 3.

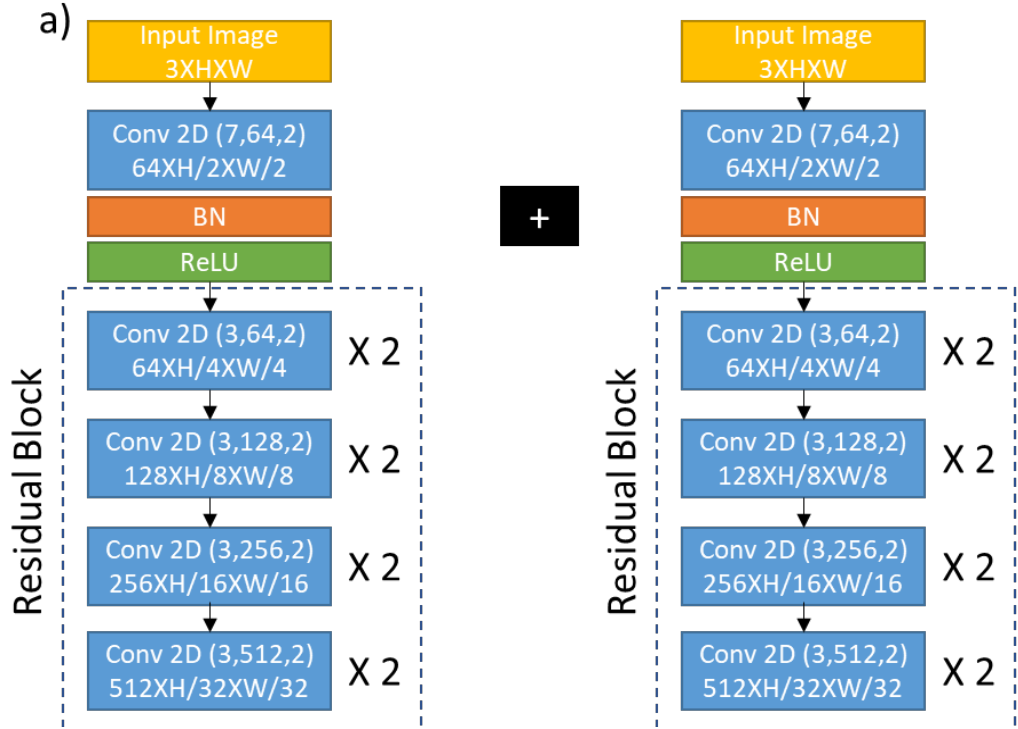

Figure 3.a) Network architecture with number of layers-Encoder

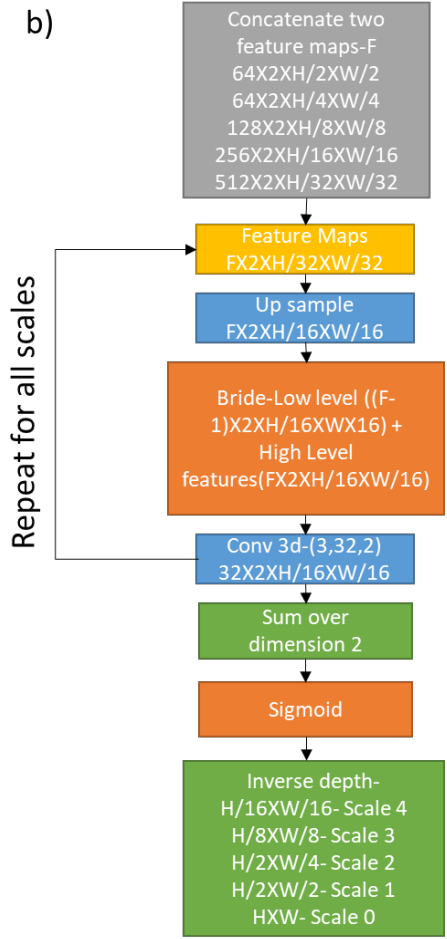

Figure 3.b) Network architecture with number of layers-Decoder

### 3.2.2 Pose Network

The network architecture of the encoder for pose network uses the same number of layers as depth encoder. The reference and two source images are given as input to the pose encoder. The feature maps are generated from the ResNet-18 encoder network. The generated features maps from reference and one source image are concatenated to form the dense layers. These feature maps are passed through pose decoder. The features are passed through four convolutional layers of 256 features each with 3×3 kernel size. These layers are concatenated together to form a feature map with 6 degrees of freedom to be predicted from a pair of images. This is passed through ReLU activation function and mean of these features are taken. These features are split into three layers for axis angle and three layers for translation. Thus, the pose is estimated from the two source images ($I_{r-1}$, $I_{r+1}$) with respect to the reference image ($I_r$) and is denoted as $P_{r'}(P_{r \to (r+1)}, P_{r \to (r-1)})$. It is composed of a 4×4 camera transformation matrix with rotation and translation parameters between source view and reference view. This pose information along with predicted depth and camera matrix is used to reconstruct the reference image as given in Equation (1).

### 3.3 Loss functions

To increase the model performance and improve the generated depth quality, various loss terms are used during training. Unlike supervised techniques which use the ground truth depth to increase the predicted depth, we use the generated reference image for backpropagation to increase the quality of generated depth. This is handled through different loss terms as described in Equation (2). The total loss is given by

$$L = \lambda_1 L_p + \lambda_2 L_S + \lambda_3 L_M + \lambda_4 L_C \qquad (2)$$

Where $\lambda_1, \lambda_2, \lambda_3, \lambda_4$ represents the weights between different loss terms used. $L_p$, represents the reprojection loss, $L_M$, represents the Masking loss and $L_C$, represents the contrastive loss term. The importance of different loss terms and the selection of weights are discussed in Section 4.

### 3.3.1 Reprojection Loss

The difference between the reconstructed reference image and the original image is minimized using the photometric loss such that the depth quality increases. This is a combination of both L1 and SSIM loss (Wang et al., 2004) for identifying the difference as shown in Equation (3).

$$L_p = \frac{1}{N}\sum \alpha \frac{(1-SSIM(I_a,I_b))}{2} + (1-\alpha)||I_a - I_b|| \quad (3)$$

Where α =0.85, $I_a$ and $I_b$ represents the reference image and the reconstructed reference image. Godard et al.,(2019) suggested the use of minimum pixel reprojection loss instead of the average reprojection loss due to its minimisation of artefacts near image borders. For our case, we tested both average and minimum reprojection loss and found that minimum reprojection loss is found to be more suitable for our objective.

### 3.3.2. Smoothness Loss

To have a smoother depth images, edge-aware smoothness loss is implemented following Wang et al.,( 2018). Due to discontinuities in depth which is reflected in the colour gradient of the image, we also used this loss to produce a normalized depth map. This is given in Equation (4)

$$L_S = |\partial_x d/d^-|e^{-|\partial_x I_r|} + |\partial_y d/d^-|e^{-|\partial_y I_r|} \quad (4)$$

Where $d/d^-$ is the normalised depth, the gradient of normalised depth is weighted with the colour gradient of the reference image $I_r$ in x and y directions.

### 3.3.3 Masking loss for dynamic objects

The dataset we use for training has a significant number of frames with dynamic objects like moving cars. Self-supervised monocular depth estimation, however, works with the assumption that only the camera is moving while the objects in the scene do not move or the camera is static. Godard et al., (2019) suggested a masking loss, in order not to consider the dynamic objects in the static scenes during loss calculation. A similar loss is implemented in our network, as our dataset has many moving cars and static scenes making it more representative of real life situation, compared to standard datasets like KITTI (Geiger et al., 2012). The loss is calculated per-pixel wise, where the reprojection error for the reconstructed image and reference image is compared with the reprojection error between reference and source images. Only those pixels where the reprojection error between the reconstructed image and the reference image is less than the reprojection error between the reference image and the source image is considered for error calculation by applying the mask, $L_M$ as given in Equation (5),

$$L_M = min[L_p(I_r, I_{r'\to r})] < min[L_p(I_r, I_{r'})] \quad (5)$$

Where $I_{r'\to r}$ represents the reconstructed image, $I_r$ represents the reference image, $I_{r'}$ represents the source image and $L_p$ represents the reprojection error.

### 3.3.4 Contrastive loss

In order to improve the quality of the generated images, we add a contrastive loss term as suggested by Spencer et al., (2020). It takes feature vectors from the reference image $I_r$ and reconstructed image $I_{r'\to r}$, calculates the distance between the two and is defined as given in Equation (6),

$$l(y, r_1, r_2) = \begin{cases} \frac{1}{2}d^2 & if\ (y=1) \\ \frac{1}{2}\{max(0, m-d)\}^2 & if\ (y=0) \\ 0 & otherwise \end{cases} \quad (6)$$

where, y indicates whether the pairs are positive or negative correspondence (1,0), d=$||r_1 - r_2||$ is the Euclidean distance calculated between the pairs, m is the margin set between positive and negative pairs. A positive pair is the one that has a distance between feature vectors low while negative pairs contribute to larger distances between the pairs. Here the pairs refer to the reference image and reconstructed reference image. This is shown in Equation (7)

$$L_C = \sum l(y, I_r, I_{r'\to r}) \quad (7)$$

The total loss is included based on whether the pairs are positive or negative or should be ignored. This loss serves in matching the features despite the appearance changes.

## 3.4 Inference

As our study is to estimate depth from monocular UAV videos, our main focus is to improve the depth estimation model. Hence, our objective is oriented towards modifying the depth network architecture. Compared to Godard et al., (2019)'s model, we added another encoder to extract more information from consecutive frames and to optimize the image information from this volumetric feature maps, a 3D depth decoder is also added. In addition to this, we included a contrastive loss term in our network to improve the image generation process through feature matching. The depth network is modified to improve the quality of the generated disparity maps and also to produce depth maps closer to the ground truth. To assess the performance of the proposed architecture, they are evaluated with the ground truth depth maps generated from Pix4d. Thus, we have not evaluated the network performance for pose estimation or modified the pose network and has adapted the available, simple architecture that suits our objective as proposed by Godard et al., (2019). The dataset used to train our model and the results from our network architecture are evaluated in the following section.

## 4. EXPERIMENTS

In this section, we introduce the dataset and then describe details of our implementation. After that, we present a quantitative and qualitative comparison. Furthermore, we conduct several ablation studies.

### 4.1 Dataset

The UAV oblique video dataset used for our work is taken from the UAVid (Lyu et al., 2020) imageries. The UAVid dataset is a combination of 42 video sequences captured from Wuhan (China) and Gronau (Germany). The original frame rate is 20 frames/second and the images are captured at a flying height of 50~100m with a flying speed of 10m/s, at an angle of 45°. The images are of size 4096×2160 and 3840×2160 for Germany and China respectively. There are 34 video sequences for China and 9 video sequences for Germany. The images from China are captured at varying height and have lots of dynamic features like cars, pedestrians, crowded streets which add extra complexity on top of the obliqueness of the view. Also, the images from China have a higher depth range, reaching to more than 500m. The training dataset from Germany in contrast, covers a smaller region, has a depth range of 150m, consists of rooftops, vegetation and is more uniform. The number of images used for training, testing and evaluation from each dataset are specified in Table 1. Some training sample images from both datasets are shown in Figure 4. The used training dataset consists of three categories, images from Germany alone, images from China alone and both combined together.

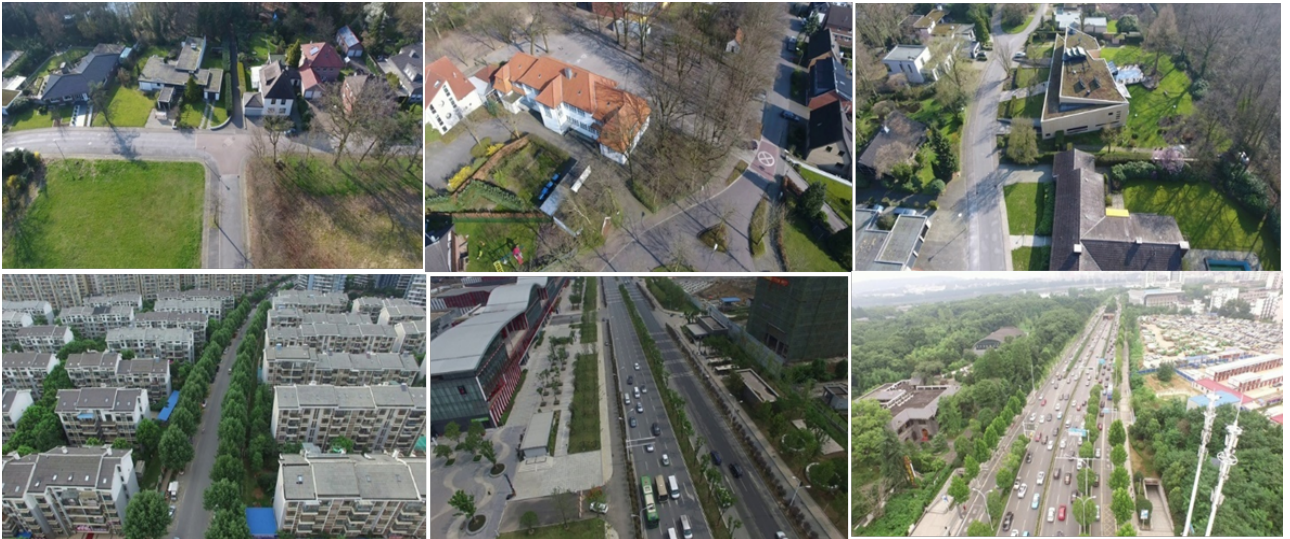

Figure 4. Training Video sequences from UAVid dataset (Lyu et al., 2020): (a) Germany (First row) and (b) China (second row).

The frame rate is an important parameter to train the model. The original frame rate of 20 frames/second, produced various noises in the results possibly due to the smaller base and parallax error. Several trials are carried out for both Germany and China dataset to determine the optimal frame rate. For Germany dataset, this is found to be 10 frames/second, while for China dataset the selection required further reduction in overlapped images. As discussed above due to a large number of high-rise buildings in Chinese dataset, the flying height had to be controlled manually. This led to varying heights and selection of a fixed frame rate due to different overlaps is complex. After several trials, it is found that there are significant noises for all frame rates due to the smaller base between many consecutive frames. In order to overcome this, frames with very narrow based are removed manually from the video sequence and then a frame rate of 1 frame/second is used for learning from the dataset.

Table 1: Dataset specification

|  | **Germany (4096×2160)** | **China (3840×2160)** |
|---|---|---|
| **Training** | 11438 | 25800 |
| **Validation** | 1500 | 2400 |
| **Testing** | 409 | 88 |

To test the model performance, the depth maps generated by the model is compared with point clouds generated using Pix4D which is considered as reference depths. The 3D reconstruction in Pix4D also encountered the problem of a narrow base between successive frames in the datasets. To widen the base and to create good quality point clouds some of the frames had to be removed which decreased the number of available reference frames. The reference depth images are theme derived from the point cloud using the P-matrix obtained from Pix4D. A total of 409 images for Germany and 88 images for China is obtained as reference depths through this process. For testing the model, a single test image is given as input to the model from which a depth map is produced which is then compared to the corresponding reference depth.

## 4.2 Reference depths

To assess the quality of the generated single depth images from various models, the results from these models are compared with ground truth depth images. For our work, the ground truth images are prepared from point clouds obtained from a commonly used Photogrammetric tool (Pix4D). The point clouds obtained from Pix4D are in real world coordinate system (WGS 1984, UTM 32N) with mean sea level as a datum. These point clouds are interpolated to form reference depth images that are compared with the test images from Germany and China dataset. To validate the model performance, the obtained inverted depths are converted to depths using an appropriate scale based on the reference depth images.

## 4.3 Implementation details

All models in this work are implemented in PyTorch (Paszke et al., 2017) with an input resolution of 640×352 pixels for both datasets. Our model is trained for 40 epochs, 20 epochs, 20 epochs each for Germany, China dataset and both datasets mixed together. We use Adam optimizer for training all three datasets. The learning rate is set to $10^{-4}$ for 75% of the epochs and $10^{-5}$ for remaining epochs. Training takes around 17, 22 and 27 hours in a single Nvidia Titan Xp GPU of 16GB memory for Germany, China and combined dataset respectively. After hyperparameter tuning, the batch size is fixed at 12. The different weights for the loss terms are tuned and the respective weights of $\lambda_1, \lambda_2, \lambda_3, \lambda_4$ are determined. The weightage of reprojection loss, auto masking loss are chosen as 1, while the weightage of smoothness loss is maintained as 0.001 and weight of contrastive loss is fixed as 0.5. The obtained disparities are converted to depths using an appropriate scale based on the model and dataset tested. In order to convert our disparities, we follow a similar scaling method as implemented by other SMDE models like Zhou et al., (2017), Repala and Dubey, (2018), Aleotti et al., (2018), Godard et al., (2019) etc. Here, the minimum and maximum depths are obtained from the reference depths which are used to fix the depth range of the model results. The obtained disparity values are then multiplied with a scaling value that produces metrically comparable results with the reference depths. From a photogrammetric perspective, the orientation from the pose networks could also be used to find the appropriate scaling for the entire image. The pose network takes an average of all the training images to determine the relative orientation parameters. However due to the added complexities of this cannot be used in direct orientation of images and to make our results comparable with other SMDE models we follow the simpler method for scaling.

## 4.4 Evaluation metrics

To assess the performance, various pixel-wise metrics are calculated between the models and reference depths. The evaluation of the accuracy is done based on calculating several metrics between the single image depths (d') generated from the model after fixing the model parameters and the reference depths (d) produced from PIX4D. This includes Absolute Relative difference (Abs Rel) given in equation (8), Squared Relative difference (Sq Rel) given in equation (9), Root Mean Square Error (RMSE) given in equation (10), accuracy given in equation (11) as described in Godard et al., (2019) and Hermann et al., (2020).

$$\text{Abs Rel} = \frac{1}{N}\sum_{i=1}^{N}\frac{|d(x_i)-d\prime(x_i)|}{d(x_i)} \quad (8)$$

$$\text{Sq Rel} = \frac{1}{N}\sum_{i=1}^{N}\frac{|d(x_i)-d\prime(x_i)|^2}{d(x_i)} \quad (9)$$

$$\text{RMSE} = \sqrt{\frac{1}{N}\sum_{i=1}^{N}(d(x_i)-d'(x_i))^2} \quad (10)$$

$$\text{Accuracy}(\delta_\theta) = \frac{1}{N}\sum_{i=1}^{N} max(\frac{d_i}{d\prime_i}, \frac{d\prime_i}{d_i}) < \theta \quad (11)$$

Here Accuracy is measured as the fraction of pixels that are within a certain threshold θ to the corresponding pixel wise value in the reference depth map. The thresholds chosen are 25%, 15% and 5% based on the standard benchmarks from KITTI quantitative assessments (Garg et al., 2016).

## 4.5 Comparison of Results

In this part, we evaluate our model results by comparing with reference depths generated from PIX4D software. Our model results are also compared with two other models, Hermann et al., (2020) and Godard et al., (2019), as they have presented a self-supervised model for aerial image depth estimation with a similar pose network and some of the loss terms that we used. All three models are trained with the same environment, dataset, resolution, batch size and number of epochs in order to make the results comparable.

**Qualitative results**

The results from various models along with reference depths are shown in Figure 5a and 5b. From the qualitative aspect, our model results over Germany showed a closer approximation to that of the reference depths. The model reconstructs depth images that smoothen out the variations over vegetation and show fine edges of buildings and roofs. It shows that our model preserves small details. The objects closer to the camera are shown as yellow and farther regions are shown in blue. The local variations in the ground surface are difficult to differentiate. Similarly, for China dataset the smooth transition from closer to farther objects is visible. But in China dataset, due to larger depth variations and reduction in image resolution, each pixel contains many objects along with the dynamic objects making this dataset more challenging to predict. From images, it is difficult to differentiate the regions which have improved as all the models are well capable of generating depths images from UAV sequences. So to accurately understand the improvement in each model quantitative evaluation is performed.

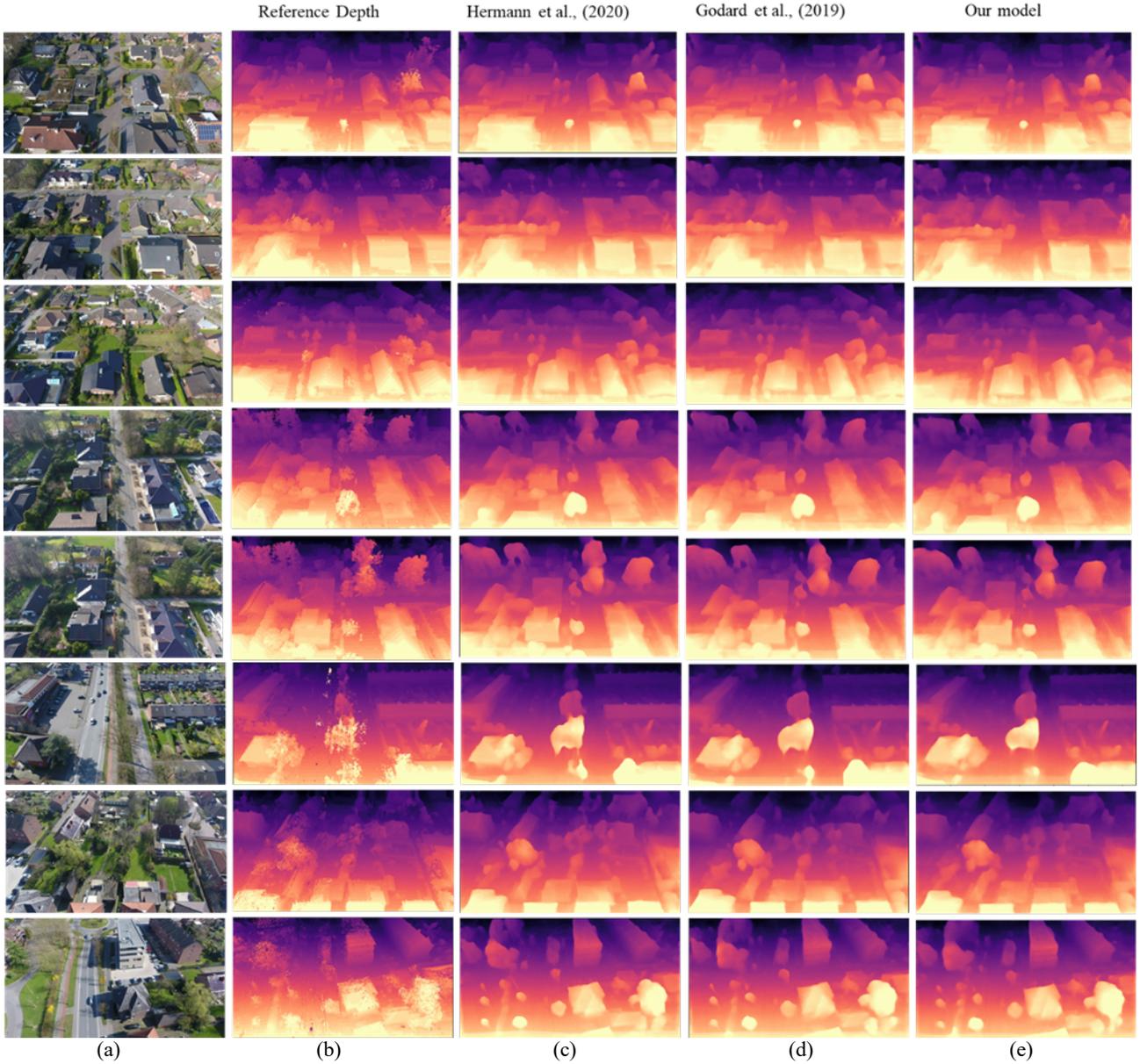

(a)          (b)          (c)          (d)          (e)

Figure 5a. Qualitative comparison between (b)Reference depths from Pix4d, (c) Hermann et al., (2020), (d) Godard et al., (2019), (e) Our model from Germany and China dataset. The test image is given in (a).

**Quantitative results**

The quantitative metrics between various models, datasets and reference depths are shown in Table 2, Table 3 and Table 4. The generalising capability of the models trained over one dataset and tested on other datasets is estimated and the corresponding error metrics are given in Table 5. The values given in the table are expressed in terms of meters, as the scaling used to convert the disparities to depths are approximated from the reference depths expressed in meters.

From the tables, we could observe that the results vary between different datasets and models. The deep learning models are sensitive to training data and that is why various studies group the datasets based on the similarity of features like rural, urban and synthetic datasets and evaluate the model performance at different regions. We wanted to evaluate how well our model can generalise over different datasets that have complex features, dynamic objects and large depth variations. This makes Germany dataset a simpler dataset to learn from due to its simplicity and lack of moving objects compared to China dataset. The metrics for Germany dataset shows that our model is well capable of generating depths from single images that are closer to reference depths. The mean absolute difference between reference depths and model results are less compared to the other two models and also the percentage of pixels that are similar are more than 98% at a threshold of 25%. Also, from Table 2, we could see our model performs better in absolute terms compared to other models.

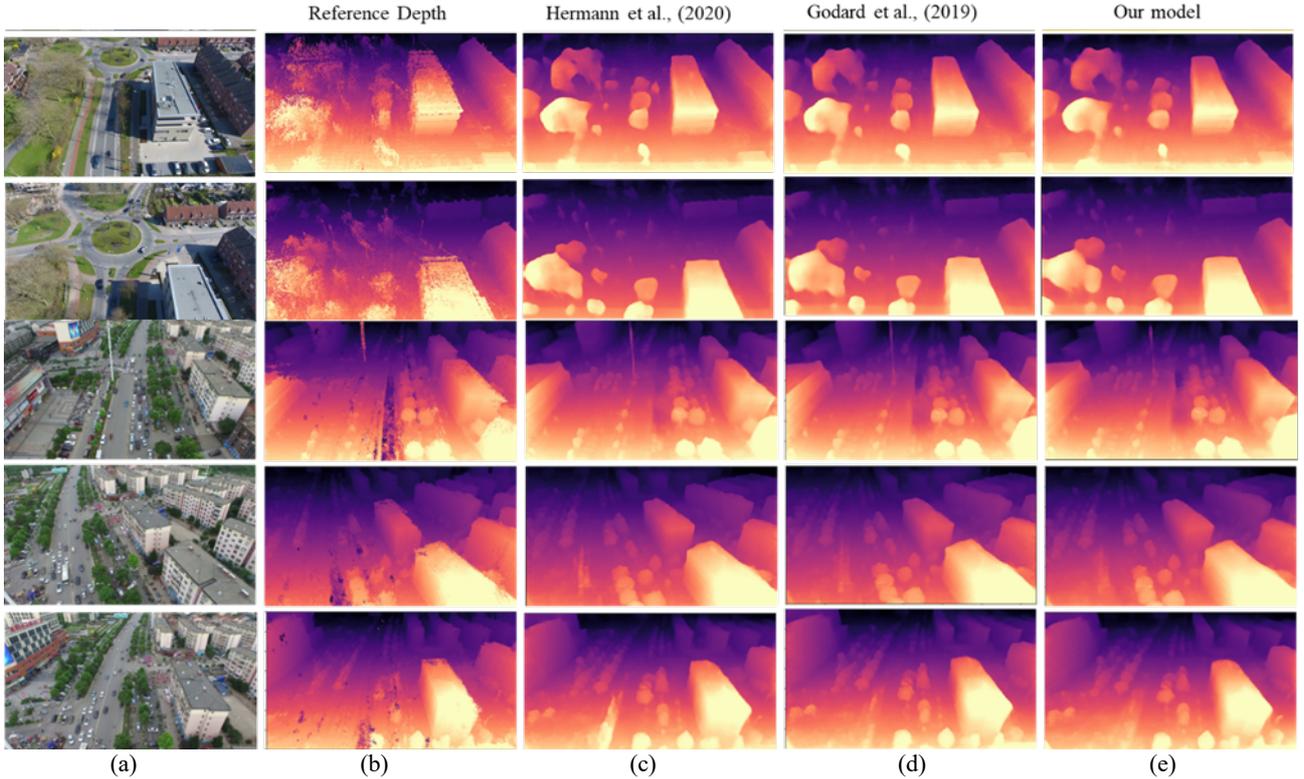

(a) (b) (c) (d) (e)

Figure 5b. Qualitative comparison between (b)Reference depths from Pix4d, (c) Hermann et al., (2020), (d) Godard et al., (2019), (e) Our model from Germany and China dataset. The test image is given in (a).

Table 2: Quantitative results achieved over different models with Germany dataset. The values represent the mean score over all the images in the corresponding test dataset.

| Method | Training dataset | Testing dataset | Abs Rel | Sq Rel | RMSE | $\delta_{1.25}$ (Higher is better) | $\delta_{1.15}$ (Higher is better) | $\delta_{1.05}$ (Higher is better) |
|---|---|---|---|---|---|---|---|---|
| Hermann et al., (2020) | Germany | Germany | 0.0372 | 0.302 | 4.275 | 0.9814 | 0.957 | 0.8085 |
| Godard et al., (2019) | Germany | Germany | 0.0371 | 0.295 | 4.1905 | 0.982 | 0.9582 | 0.810 |
| Our model | Germany | Germany | **0.0366** | **0.2925** | **4.1735** | **0.982** | **0.9582** | **0.810** |

Table 3: Quantitative results achieved over different models with China dataset. The values represent the mean score over all the images in the corresponding test dataset.

| Method | Training dataset | Testing dataset | Abs Rel | Sq Rel | RMSE | $\delta_{1.25}$ (Higher is better) | $\delta_{1.15}$ (Higher is better) | $\delta_{1.05}$ (Higher is better) |
|---|---|---|---|---|---|---|---|---|
| Hermann et al., (2020) | China | China | 0.109 | 7.858 | 49.828 | 0.874 | 0.755 | 0.337 |
| Godard et al., (2019) | China | China | **0.105** | 7.668 | 48.864 | **0.879** | **0.764** | **0.376** |
| Our model | China | China | 0.109 | **7.742** | **48.303** | 0.878 | 0.761 | 0.327 |

In higher depth ranges, like in China dataset which is used for training and testing, we could see from Table 3 that the RMSE values are larger than that of Germany dataset. For regions captured in China, farther points in oblique images might introduce errors, possibly due to reduction in image resolution resulting in many objects like trees, road, cars appear in just a single pixel. This along with the dynamic objects, moving cars and pedestrians, makes the process of learning highly challenging. Also, the maximum depth range in Germany dataset would be 150m while for China it is more than 500m. The inverse depths or the disparities generated from China datasets has pixel values less than 0.1. This means the pixel shift between images is less than 0.1 making it difficult for estimating depth. Increasing the base between images did not reduce this problem as it introduced errors due to occlusions between objects in image reconstruction task due to the larger shift. Several experiments have been carried out to find an appropriate base that reduces the impact of occlusions while having sufficient base for 3D reconstruction. Even after these experiments, there is the presence of pixels with very less disparity values (lesser than 0.1), providing more uncertainty for farther objects. To further evaluate this, instead of using the entire image, we used only pixels with depths less than 200m for evaluation of error metrics. The obtained RMSE values are reduced by more than half the current numbers presented in Table 3. Whereas all models are having difficulty in learning from the China dataset, in comparative terms our model still performs better than the other two models. It achieved an accuracy of almost 88% at a threshold level of 25%.

Table 4: Quantitative results achieved over different models with both combined datasets. The values represent the mean score over all the images in the corresponding test dataset.

| Method | Training dataset | Testing dataset | Abs Rel | Sq Rel | RMSE | $\delta_{1.25}$ (Higher is better) | $\delta_{1.15}$ (Higher is better) | $\delta_{1.05}$ (Higher is better) |
|---|---|---|---|---|---|---|---|---|
| Hermann et al., (2020) | Germany + China | Germany | 0.0391 | 0.3181 | 4.42 | 0.981 | 0.956 | 0.786 |
| Godard et al., (2019) | Germany + China | Germany | 0.0392 | 0.3147 | 4.3722 | 0.981 | 0.955 | 0.786 |
| Our model | Germany + China | Germany | **0.0381** | **0.322** | **4.364** | **0.982** | **0.956** | **0.797** |
| Hermann et al., (2020) | Germany + China | China | 0.133 | 13.390 | 63.730 | 0.836 | 0.673 | 0.215 |
| Godard et al., (2019) | Germany + China | China | 0.132 | 11.840 | 60.631 | 0.839 | 0.670 | 0.225 |
| Our model | Germany + China | China | **0.117** | **9.664** | **54.855** | **0.859** | **0.734** | **0.302** |

Table 5: Quantitative results achieved over different models trained over one dataset and tested on other to find the transferability potential of the model. The values represent the mean score over all the images in the corresponding test dataset.

| Method | Training dataset | Testing dataset | Abs Rel | Sq Rel | RMSE | $\delta_{1.25}$ (Higher is better) | $\delta_{1.15}$ (Higher is better) | $\delta_{1.05}$ (Higher is better) |
|---|---|---|---|---|---|---|---|---|
| Hermann et al., (2020) | Germany | China | 0.217 | 12.532 | 56.049 | 0.631 | 0.370 | 0.116 |
| Godard et al., (2019) | Germany | China | 0.215 | **12.377** | 55.832 | 0.636 | 0.376 | 0.117 |
| Our model | Germany | China | **0.214** | 12.417 | **55.048** | **0.645** | **0.379** | **0.118** |
| Hermann et al., (2020) | China | Germany | 0.2054 | 4.231 | 17.3812 | 0.5076 | 0.2894 | 0.0936 |
| Godard et al., (2019) | China | Germany | 0.2016 | 3.766 | 16.17 | 0.523 | 0.2994 | 0.0962 |
| Our model | China | Germany | **0.19** | **3.0494** | **14.4122** | **0.5548** | **0.3298** | **0.11** |

Finally, we combined images from both Germany and China dataset to understand how well the model performs when it has a diverse range of images during training. To evaluate the performance, the obtained model was tested on Germany and China dataset separately. From Table 4, we could see that the model performs comparatively better with Germany dataset than China dataset. Also, we can see that training with individual datasets produces better results than combining them. Here also we could observe the pattern of our model performing better than the other two models. The difference between our model with other models is higher in China dataset compared to Germany dataset. This could also show that as the depth variations increase and the difficulty in handling farther objects are handled slightly better by our model in comparison to the other two.

In order to test the model transferability and its ability to generalise, we trained the model with images from one dataset and tested it on the other. From Table 5, we can see that the model performance is reduced, still in all models are capable of generating depth maps which are trained in different depth ranges. This is mainly due to the exposure of model with one particular landscape which caused the reduction in various metrics. Our model requires only images for training due to its self-supervised nature, so training in a different region can be handled through training with new image sequences for improvement or our model can be fine-tuned on other image sequence based on the requirement.

## 4.5 Ablation study

To study the impact of various loss terms and the image resolution, we carried out various experiments to understand the model performance. For quantitative evaluation, we used only Germany dataset for these experiments as they showed closer approximation to the ground truth than other datasets.

**Scratch:** Our model uses pre-trained encoders for the training process. In order to understand the effect of pre-trained models, we experimented by training our encoders from scratch. From Table 6, we could see that pre-training improves the results and there is a reduction in an absolute difference between reference depths and our model result. Also, training a model with pre-trained weights might help the model in finding the global minima easier compared to training from scratch.

**Input resolution:** As we could see from Table 6, input resolution has a large impact on the results due to mixed features within a pixel. Due to the limitation of the graphics card, we opted 640×352 resolution for our model. UAV images are of high resolution and reducing the images to lower size introduced uncertainties in results. To further see this, we also reduced the image resolution to half the resolution from our model size (320X192 pixels) and trained the model. The reduction in model generation capability and the errors caused due to the smoothening of pixels are seen in Figure 6.

**Loss terms:** To further assess the effects of loss terms on our model, we tested by removing and adding different loss terms and different combinations. As we could see from Table 6, our proposed model with all four loss terms performed better than all the other combinations. It is understood that each loss contributes a little to model improvement. Adding contrastive loss terms without masking loss also shows closer performance to our proposed model. The use of contrastive loss for improving the image generation is clearly visible from these metrics. The RMSE values between reference depth and model generated results shows a reduction in values when the contrastive loss is included. In Table 6, we could observe that the absolute values for all metrics between models without contrastive loss and after using contrastive loss showed an improvement.

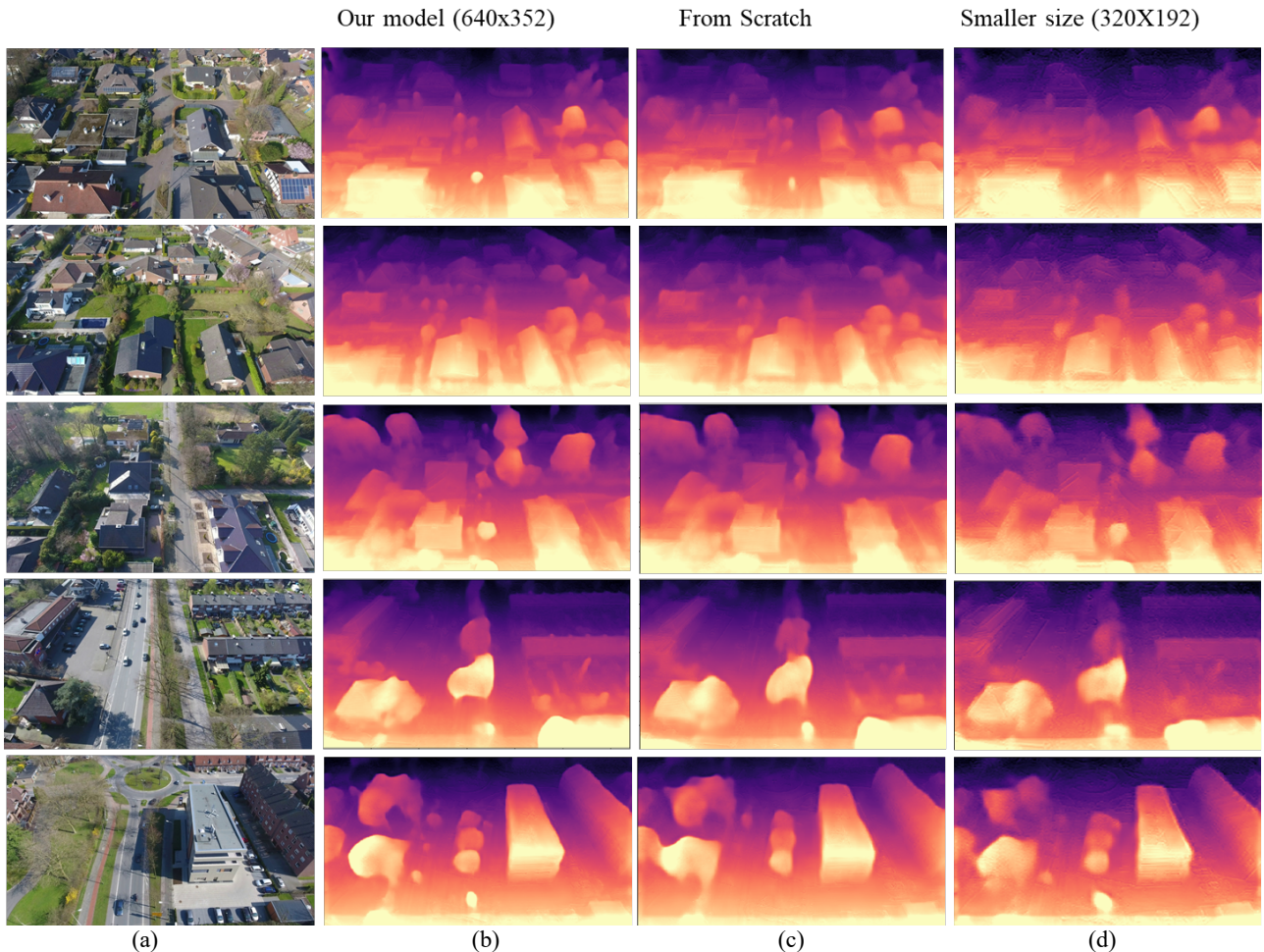

Figure 6. Qualitative results: (b) Our model, (c) Model trained from scratch, (d) Model trained with smaller resolution. The test image is given in (a).

Table 6: Quantitative results with different experiments to our model trained and tested over Germany dataset. The values represent the mean score over all the images in the corresponding test dataset.

| Method | Abs Rel | Sq Rel | RMSE | δ$_{1.25}$ (Higher is better) | δ$_{1.15}$ (Higher is better) | δ$_{1.05}$ (Higher is better) |
|---|---|---|---|---|---|---|
| Baseline- pretrained with four loss (640×352) | **0.0366** | **0.2925** | **4.1735** | **0.982** | **0.9582** | 0.810 |
| Baseline- Scratch with four loss | 0.0391 | 0.3091 | 4.287 | 0.9821 | 0.956 | 0.7867 |
| Baseline- pretrained with four loss smaller resolution (320×192) | 0.0481 | 0.3606 | 4.771 | 0.982 | 0.948 | 0.662 |
| Baseline- pretrained with three loss (reprojection, smoothness and contrastive loss) | 0.0366 | 0.294 | 4.195 | 0.982 | 0.958 | **0.811** |
| Baseline- pretrained with three loss (reprojection, smoothness and Masking loss) | 0.0372 | 0.3014 | 4.234 | 0.981 | 0.957 | 0.8087 |
| Baseline- pretrained with two loss (reprojection and smoothness loss) | 0.0366 | 0.294 | 4.201 | 0.982 | 0.957 | 0.809 |

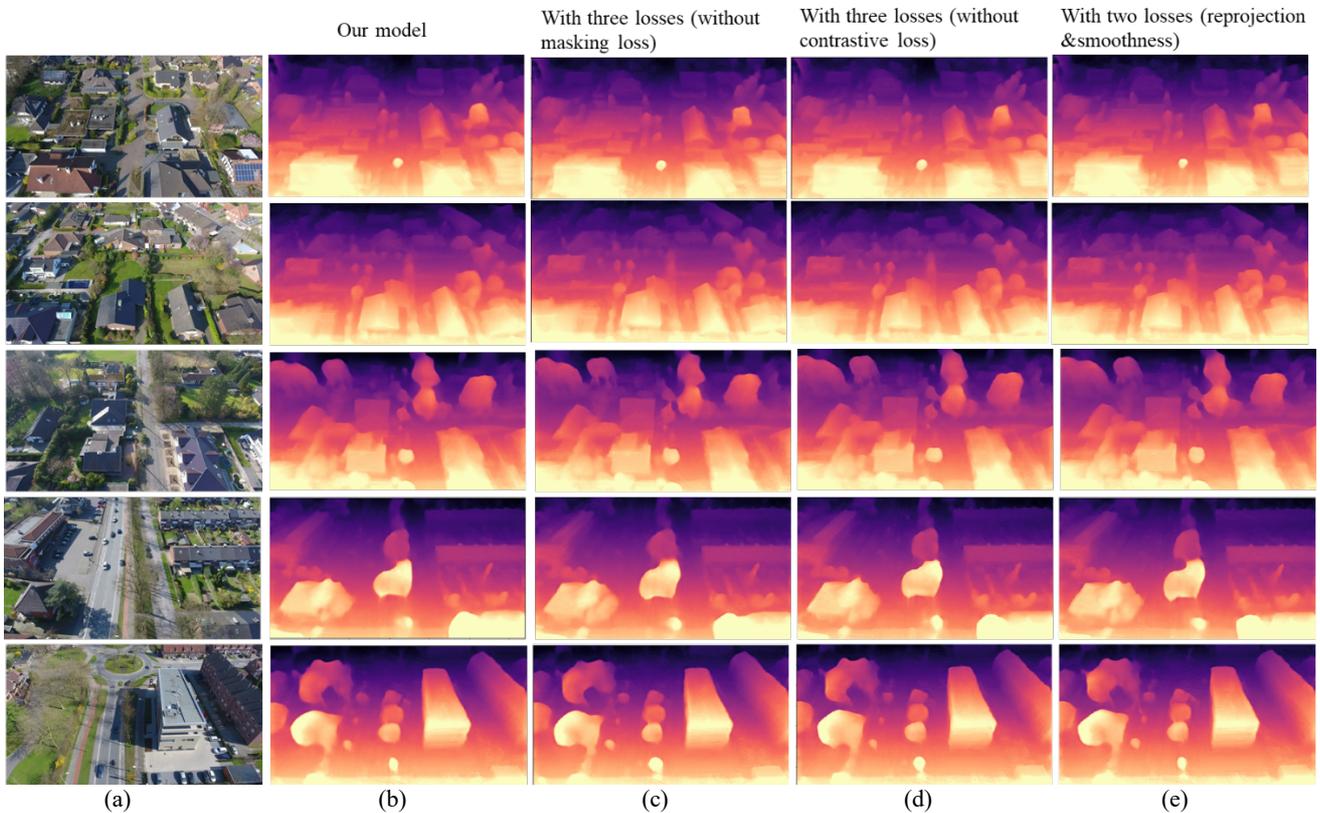

(a)     (b)     (c)     (d)     (e)

Figure 7. Qualitative results: (b) Our model, (c) Model trained with three losses (without masking loss), (d) Model trained with three losses(without contrastive loss), (e) Model trained with two losses (reprojection & smoothness loss). The test image is given in (a).

## 5. CONCULSION

This paper presents a novel approach for estimating depth information from an oblique UAV video. Our model is based on a self-supervised approach that does not require ground truth depths for training and can be trained with a UAV video sequence captured over different regions. Our model learns both depth and pose information using two networks during the training stage. The depth network consists of two encoders to capture as much image information from consecutive images and pose network calculates the relative camera position from three images. Only depth network is used for the inference stage to predict the depths from single UAV images reducing the processing time significantly. The contrastive loss term is added to improve the image generation and to increase the similarity between reconstructed images. Our model performs the best over Germany subset in UAVid video dataset due to its smaller depth range and less complex scene. The predicted depths from our model compare well with the reference depths and are also better than other state-of-the-art methods. For future work, we would like to explore further the depth estimation model for very large depth ranges in a scene along with the dynamic objects.